\pdfoutput=1

\documentclass[11pt]{article}

\usepackage[final]{acl}

\usepackage{times}
\usepackage{latexsym}

\usepackage[T1]{fontenc}

\usepackage[utf8]{inputenc}

\usepackage{microtype}

\usepackage{inconsolata}

\usepackage{hyperref}       
\usepackage{url}            
\usepackage{booktabs}       
\usepackage{amsfonts}       
\usepackage{nicefrac}       
\usepackage{microtype}      
\usepackage{xcolor} 
\usepackage{natbib}
\usepackage{adjustbox}
\usepackage{amsmath}
\usepackage{subcaption}
\usepackage{appendix} 
\usepackage{algorithm}
\usepackage{algpseudocode}

\usepackage{multirow}

\usepackage{soul}

\newcommand{\seq}[1]{\boldsymbol{#1}}
\usepackage{tabularx}
\usepackage{pifont}
\newcommand{\cmark}{\ding{51}}%
\newcommand{\xmark}{\ding{55}}%

\title{Retrieval-Augmented Multi-Modal Chain-of-Thoughts Reasoning for Large Language Models}

\author{%
  Bingshuai Liu\thanks{These authors contributed equally to this work.} \\
  Xiamen University\\
  \texttt{bsliu@stu.xmu.edu.cn}  \\
  \And
  Chenyang Lyu\footnotemark[1] \\
  MBZUAI \\
  \texttt{chenyang.lyu@mbzuai.ac.ae}  \\
  \And
  Zijun Min \\
  Xiamen University\\
  \texttt{minzijun@stu.xmu.edu.cn} \\
  \AND
   Zhanyu Wang\\
  Tencent AI Lab\\
  \texttt{zhanyuwang@tencent.com} \\
  \And
  Jinsong Su\thanks{Corresponding author.} \\
  Xiamen University\\
  \texttt{jssu@xmu.edu.cn}\\
    \And
  Longyue Wang \\
  Tencent AI Lab\\
  \texttt{vinnylywang@tencent.com} 
}

\begin{document}
\maketitle
\begin{abstract}
The advancement of Large Language Models~(LLMs) has brought substantial attention to the Chain of Thought~(CoT) approach, primarily due to its ability to enhance the capability of LLMs on complex reasoning tasks. Moreover, the significance of CoT approaches extends to the application of LLMs for multi-modal tasks. However, the selection of optimal CoT demonstration examples in multi-modal reasoning remains less explored for LLMs due to the inherent complexity of multi-modal examples. In this paper, we introduce a novel approach that addresses this challenge by using retrieval mechanisms to dynamically and automatically select demonstration examples based on cross-modal and intra-modal similarities. Furthermore, we employ a Stratified Sampling method of categorising demonstration examples into groups based on their types and then retrieving examples from different groups respectively to promote the diversity of demonstration examples. Through a series of experiments on two popular benchmark datasets: ScienceQA and MathVista, we demonstrate that our approach significantly improves the performance of GPT-4 by 6\% on ScienceQA and 12.9\% on MathVista, and enhances the performance of GPT-4V on two datasets by 2.7\%, substantially improving the performance of the most advanced LLMs and LMMs for complex multi-modal reasoning tasks.

\end{abstract}

\section{Introduction}

\begin{figure}[!ht]
    \centering
    \includegraphics[width=1.0\linewidth]{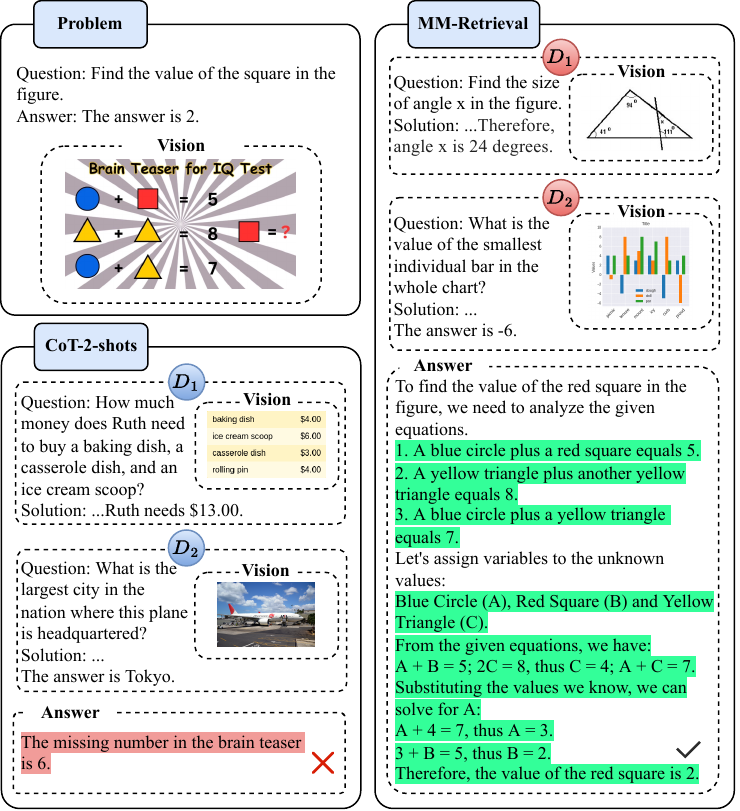}
    \caption{Our MM-Retrieval approach dynamically retrieves demonstrations based on the question. Compared with CoT, it has better adaptability and can stimulate the reasoning ability of LLMs. The red $D_1$, $D_2$ represent demonstrations retrieved based on the question, while the blue $D1$, $D2$ represent the fixed samples regardless of the question.}
    \label{fig:motivation}
\end{figure}

The field of Natural Language Processing (NLP) has experienced significant advancements due to the emergence of Large Language Models (LLMs), which have reshaped the landscape of many tasks for their extensive capabilities. A key technique that has contributed greatly to the success of LLMs is the Chain of Thought (CoT) technique, as documented in prior studies~\citep{wei2022chain,kojima2022large}. This technique becomes particularly crucial, especially when applied to multi-modal tasks. One of the most prominent applications is multi-modal question answering, which involves reasoning with both text and 
images~\citep{zhang2023multicot,lu2023chameleon,lyu2023macaw,li2023comprehensive}. However, as researchers delve further into the integration of CoT with LLMs~\citep{wang2022self-consistency,least,autocot}, the selection of appropriate demonstration examples to guide multi-modal reasoning emerges as a recurring challenge. Given that multi-modal examples often combine the complexities of both text and visual data, identifying the most relevant and informative examples is a non-trivial task~\citep{bar2022visual_icl,li2023otter_mm_icl,li2023mimicit_mm_icl}.

\begin{figure*}[!]
    \centering
    \includegraphics[width=0.7\linewidth]{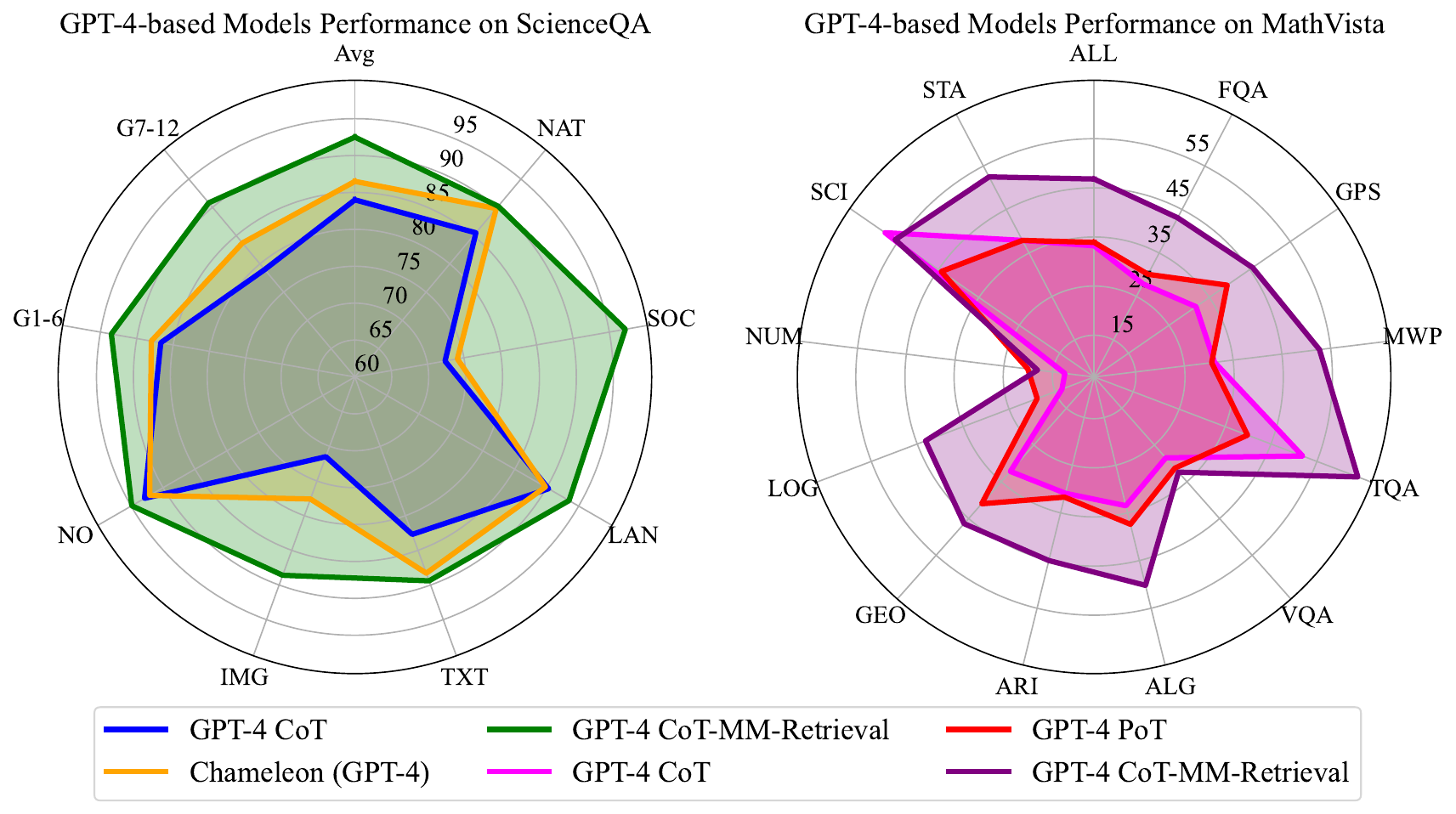}
    \caption{Results on different categories of ScienceQA~\citep{lu2022learn} and MathVista~\citep{lu2023mathvista}. Our proposed approach obtains substantial improvements over previous baseline models including \textit{CoT}~\citep{lu2023chameleon}, \textit{PoT}~\citep{lu2023mathvista} and \textit{Chameleon}~\citep{lu2023chameleon} on GPT-4 foundation models.}
    \label{fig:radar-chart}
\end{figure*}

\begin{figure*}[t]
  \centering
    \includegraphics[width=0.8\textwidth]{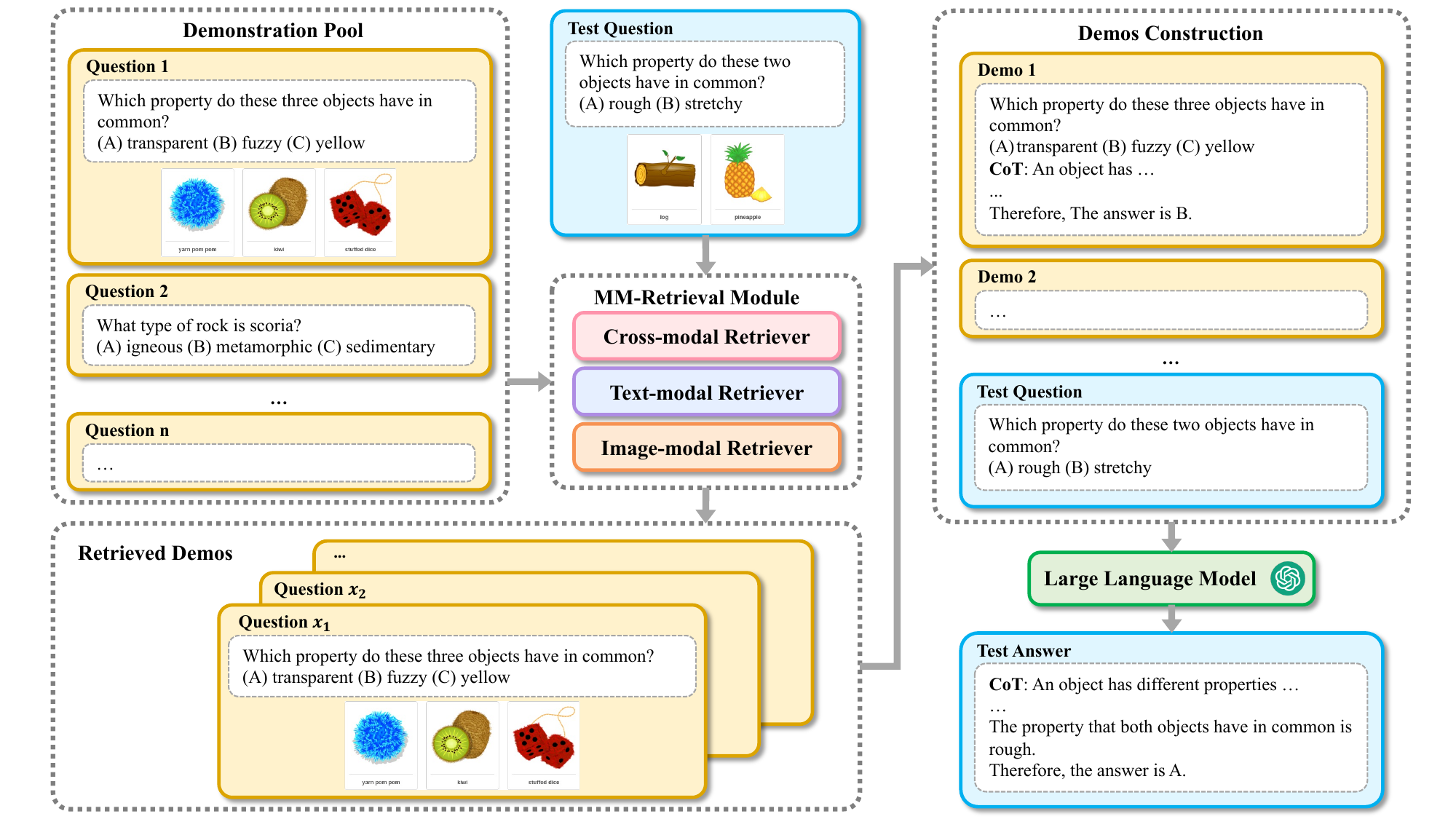}
  \caption{An overview of our proposed multi-modal retrieval method. We employ both cross-modality retrieval and intra-modality retrieval (text-modal and image-modal retrieval), to obtain relevant examples as retrieved demonstrations from demonstration pool. Then, these retrieved demonstrations are integrated with prompt and test question, serving as the input for LLMs.}
  \label{fig:mm-retrieval}
\end{figure*}

To address this issue, our paper presents a new method that utilizes retrieval mechanisms to select demonstration examples both dynamically and automatically. As shown in Figure~\ref{fig:motivation}, relevant and informative demonstration examples retrieved by our approach can elicit the reasoning ability of LLMs and lead to the correct answer. An overview of our proposed approach is shown in Figure~\ref{fig:mm-retrieval}. Our approach primarily focuses on utilizing cross-modal similarity and intra-modal similarity. We leverage these similarities to retrieve demonstration examples with the aim of enhancing the CoT reasoning process with more relevant demonstration examples on multi-modal tasks~\citep{zhang2023makes_visual_icl,sun2023exploring_visual_icl}. To further ensure a more comprehensive and diverse selection, we innovatively utilize Stratified Sampling~\citep{pmlr-v48-liberty16}. This simple but effective method involves sampling in sequence according to the group to which the retrieval sample belongs. By sampling from different groups, we aim to provide LLMs with a diverse range of examples, thus improving the overall quality of multi-modal reasoning.

To evaluate the effectiveness of our proposed approach, we performed extensive experiments on two benchmark multi-modal QA datasets, ScienceQA~\citep{lu2022learn} and MathVista~\cite{lu2023mathvista}. These experiments demonstrate that our approach substantially enhances the performance of LLMs, establishing a new state-of-the-art in multi-modal reasoning tasks. On the ScienceQA and MathVista dataset, our approach has shown substantial improvements, as illustrated in Figure~\ref{fig:radar-chart}. 
For the ScienceQA dataset, the ChatGPT-based and GPT-4-based retrieval methods outperform the state-of-the-art Chameleon by 4.8\% and 4.4\% respectively. With more demonstrations, the optimal performance of ChatGPT-based and GPT-4-based methods can reach 86.4\% and 92.5\%. For GPT-4V, our method can achieve an average accuracy improvement of 2.7\% compared to the zero-shot setting.

Furthermore, our approach also demonstrates superior performance on MathVista dataset. Our approach based on ChatGPT and GPT-4 obtains substantial improvements of 8.4\% and 13.6\% respectively. Moreover, our approach can further boost the performance of the most advanced LLM - GPT-4V, improving its overall accuracy on MathVista by 2.7\%, which demonstrates the effectiveness of our approach. 

We also conducted comprehensive experiments on the contributions of each part of our method, including the visual information, retrieval mechanism, and Stratified Sampling. Additionally, we conducted detailed analysis to study the influence of varying the number of demonstration examples, providing valuable insights of how our proposed approach works with LLMs for multi-modal tasks.

\section{Related Work}

\subsection{Retrieval-Augmented Generation for LLMs}

Retrieval-augmented generation (RAG) for LLMs represents an important advancement in enhancing the generative capabilities of models by integrating external knowledge sources. Early works such as REALM \cite{guu2020realm} and RAG \cite{lewis2020retrieval} introduced the foundational methodology for incorporating external documents into the generation process. Subsequent research expanded the retrieval-augmented paradigm to multi-modal contexts. Such as MuRAG \cite{chen2022murag} and REVEAL \cite{hu2022reveal}, which augment language generation with both textual and visual information from external sources. Furthermore, recent studies such as FiD-Light \cite{hofstatter2022fid} and REPLUG \cite{shi2023replug} have focused on improving the efficiency and effectiveness of retrieval-augmented systems as well as exploring in-context retrieval-augmented mechanisms \cite{ram2023incontext,de2023lumen}.

\subsection{In-Context Learning}
In-Context Learning (ICL) uses LLMs for tasks based on a handful of examples embedded within the context~\citep{bert,gpt2,gpt3,chowdhery2022palm}, showing effectiveness in NLP and complex mathematical reasoning~\citep{cot}. The setup for ICL involves generating responses from LLMs using context provided by task guidance and demonstration examples. It's sensitive to prompt structure, example selection, and example order~\citep{calibrate,wang-etal-2023-towards-cot,fu2022complexitycot}. The application of ICL extended to multi-modal tasks, including image segmentation, synthesis, and text-to-speech synthesis~\citep{bar2022visual_icl, wang2023imagesPainter, wang2023seggpt, wang2023prompt_diffusion, tsimpoukelli2021frozen, alayrac2022flamingo}, with potential in structured spaces like graphs~\citep{huang2023graph_icl}.

\subsection{Chain-of-Thought Reasoning}
Chain-of-Thought (CoT) reasoning guides LLMs to step-by-step reason, improving performance across arithmetic, symbolic, and logic tasks~\citep{cot,kojima2022large}. Approaches included sampling multiple reasoning paths~\cite{wang2022self-consistency}, dividing complex problems into sub-problems~\cite{least}, and dynamically selecting examples for few-shot prompting~\cite{autocot,shi2022language}. In addition to textual data, CoT has been adapted for tabular data~\cite{ziqi-lu-2023-tab-cot} as well. Furthermore, its potential has been explored in multi-modal settings, demonstrating enhanced reasoning capabilities through the fusion of language and vision~\cite{zhang2023multicot,lu2023chameleon,lu2023multimodal}. Studies like~\cite{zhang2023multicot} proposed a two-stage CoT framework for multi-modal tasks, significantly improving reasoning accuracies on benchmarks like ScienceQA. Chameleon~\cite{lu2023chameleon} introduced plug-and-play modules for Large Multi-modal Models~(LMMs), enabling complex reasoning by combining different tools.

\section{Methodology}

\begin{figure}[!]
    \centering
    \includegraphics[width=1.0\linewidth]{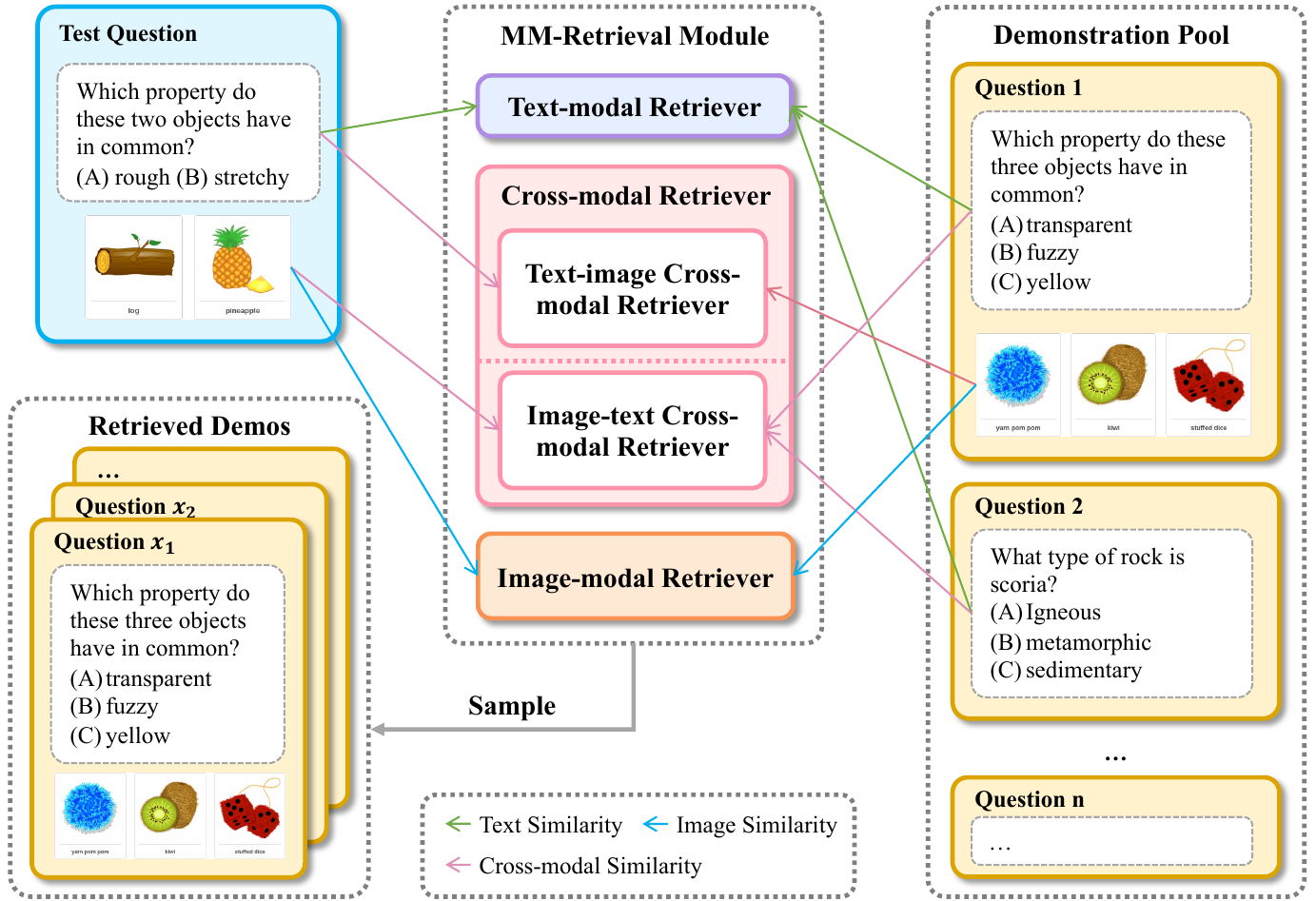}
    \caption{A detailed illustration of our multi-modal retrieval approach, where we use intra-modal similarity and cross-modal similarity to sample demonstration examples $\seq{D}$ from demonstration pool $ \seq{Q} $.}
  \label{fig:retrieve_details}
\end{figure}

Our methodology is anchored in the CoT in-context learning paradigm, which effectively leverages LLMs. For every input query, we aim to harness a set of relevant CoT demonstration examples to booster the LLM's reasoning abilities. To achieve this, we introduce a novel approach that employs retrieval mechanisms for the dynamic and automatic selection of demonstration examples, as well as incorporating visual knowledge into the prompt. A detailed illustration of our approach is shown in Figure~\ref{fig:retrieve_details}. Central to our methodology is the extraction of cross-modal similarity and intra-modal similarity, crosswise between textual~$\seq{q}^{t}$ and visual context~$\seq{q}^{v}$ of the test question~$\seq{q}$ and examples in demonstration pool~$\seq{Q}=\{\seq{q}_{1},......,\seq{q}_{n}\}$.
Another distinctive feature of our methodology is the incorporation of Stratified Sampling. By categorizing demonstration examples into distinguishable groups according to their inherent attributes, we aim to expand the diversity of selected examples. Retrieving from diverse groups ensures that the LLMs receive a multifaceted set of demonstrations, ultimately enhancing the robustness of multi-modal reasoning.

\subsection{Incorporation of Visual Information to LLMs}
Our method works for both LLMs and LMMs, and our task is multi-modal QA task which contains image and corresponding text question. It poses a significant challenge for LLMs to answer correctly without transitioning the image modality into the textual modality through an auxiliary visual expert model. Therefore, it's very important for LLM to acquire the visual information of the question through visual information model.
Following the implementation of Chameleon and MathVista, our visual information model mainly includes two parts:

\paragraph{Image Captioning}
We employ an image captioning model to obtain textual description of the given image. The image captioning result is represented as: \{$V_{c}$\}, which is a segment of text expressing the main content of the image.

\paragraph{Optical Character Recognition}
Besides image captioning system, we also use the Optical Character Recognition~(OCR) system to recognize textual characters in the given image, and the detected texts are represented as \{$V_{o}$\}.

So, the visual information we used is represented as $\seq{V}=\{V_{c}, V_{o}\}$, which is the concatenation of the generated image caption and texts detected by OCR systems.

\subsection{Retrieval Mechanism}
Suppose we have a test example $\seq{q}$ to be answered, which consists of a visual context $\seq{q}^{v}$~(usually an image) and a textual context $\seq{q}^{t}$~(usually the description of the question). Each question in $\seq{Q}$ has the same components as $\seq{q}$, so $\seq{q}_{i}=\{\seq{q}^{v}_{i}, \seq{q}^{t}_{i}\}$ for $\seq{q}_{i} \in \seq{Q}$. Meanwhile, we also have a multi-modal question collection represented as $\seq{Q}=\{\seq{q}_{1},......,\seq{q}_{n}\}$, from where we can gather demonstration examples to help LLM answer the test example $\seq{q}$.  Using a retrieval function, demonstrations are extracted from $\seq{Q}$ to form the retrieved demonstration set $\seq{D}$. The general retrieval process can be expressed as:

\begin{align}
\seq{D} = F_{k}(\seq{q}, \seq{Q}) = \underset{\seq{q}_{i} \in \seq{Q}}{\mathrm{arg\,max}^k} \left( \frac{F_{e}(\seq{q}) \cdot F_{e}(\seq{q}_{i})^{T}}{\|F_{e}(\seq{q})\| \|F_{e}(\seq{q}_{i})\|} \right)
\label{eq:general_retrieval}
\end{align}
where $F_{\seq{e}}$ denotes our encoder model for encoding $\seq{q}$, $F_{e}(\seq{q}) \in \mathbf{R}^{1\times h}$ and $F_{e}(\seq{q}_{i}) \in \mathbf{R}^{1\times h}$. $k$ indicates that we sample the top-k examples from $\seq{Q}$ that maximize the cosine similarity with $\seq{q}$. Then the sampled top-k examples serve as the demonstration examples.  

Specifically, considering the complexity of retrieval in multi-modal settings, we extend Equation~\ref{eq:general_retrieval} by employing $\seq{q}^{v}$ and $\seq{q}^{t}$ to retrieve demonstration examples from $\seq{Q}$ respectively:

\begin{equation}
\begin{split}
D = F_{k}(\mathbf{q}, \mathbf{Q}) = F_{k_{1}}(\mathbf{q}^{v}, \mathbf{Q}^{v}) &\cup 
 F_{k_{2}}(\mathbf{q}^{t}, \mathbf{Q}^{t}) \cup \\
 F_{k_{3}}(\mathbf{q}^{v}, \mathbf{Q}^{t}) &\cup 
 F_{k_{4}}(\mathbf{q}^{t}, \mathbf{Q}^{v})
\end{split}
\label{eq:mm_retrieval}
\end{equation}

where $F_{k_{1}}(\seq{q}^{v}, \seq{Q}^{v})$ means we retrieve the top-$k_{1}$ demonstration examples from $\seq{Q}$ based on the cosine similarity between $F_{e}(\seq{q}^{v})\in \mathbf{R}^{1\times h}$ and $F_{e}(\seq{q}^{v}_{i})\in \mathbf{R}^{1\times h}$, which represents comparison between the visual context of $\seq{q}$ and example $\seq{q}_{i}$ in the demonstration pool, the same applies for the others. It's worth noting that $k=\sum_{i=1}^{4}k_{i}$. The first two terms in the right side of Equation~\ref{eq:mm_retrieval} represent retrieval based on intra-modal similarity whereas the latter two terms indicate cross-modal retrieval. $F_{e}$ can be any appropriate encoders to obtain embeddings for $\seq{q}^{t}$~(\textsc{Text-Encoder}) and $\seq{q}^{v}$~(\textsc{Visual-Encoder}).

\subsection{Sampling Method}

Furthermore, to maintain diversity and relevance in our demonstration examples, we employ the approach of Stratified Sampling~\citep{pmlr-v48-liberty16}. This approach allows us to sample demonstration examples from the four groups that we retrieved based on cross-modal and intra-modal similarity. For the sake of simplicity, we simplify $F_{k_{1}}(\mathbf{q}^{v}, \mathbf{Q}^{v})$ to $D_{k_1}^{I2I}$, $F_{k_{2}}(\mathbf{q}^{t}, \mathbf{Q}^{t})$ to $D_{k_2}^{T2T}$, $F_{k_{3}}(\mathbf{q}^{v}, \mathbf{Q}^{t})$ to $D_{k_3}^{I2T}$ and $F_{k_{4}}(\mathbf{q}^{t}, \mathbf{Q}^{v})$ to $D_{k_4}^{T2I}$.
The Stratified Sampling process $S$ is as follows:

\begin{equation}
\begin{split}
D &= S(D_{k_1}^{I2I}, D_{k_2}^{T2T}, D_{k_3}^{I2T}, D_{k_4}^{T2I}) \\
&= \{d_{i}^{k_1}, d_{i}^{k_2}, d_{i}^{k_3}, d_{i}^{k_4}\}_{i=1}^{n}
\end{split}
\label{eq:mm_retrieval_sampling}
\end{equation}

where $d_{i}^{k1} \in D_{k_1}^{I2I}$, $d_{i}^{k2} \in D_{k_2}^{T2T}$, $d_{i}^{k3} \in D_{k_3}^{I2T}$ and $d_{i}^{k4} \in D_{k_4}^{T2I}$. Moreover, to accommodate the different complex characteristics of multi-modal data~(e.g. examples in different domain with various nature), we propose the adaptive use of Stratified Sampling when dealing with different types of questions $\seq{q}$.  Specifically, we use an empirical function $\sigma$ to decide whether to adopt Stratified sampling or not~(i.e. $\sigma$ can be the performance $\Delta$ on dev set or other heuristics). The detailed sampling strategy is shown in Table~\ref{tab_sampling_strategy}.

\subsection{Final Prediction}

With the retrieved demonstration examples, our next step is to predict the final answer for the given test question $\seq{q}$. To achieve this, we combine the test question $\seq{q}$ with the retrieved demonstration set $\seq{D}$ as well as the visual information $\seq{V}$. The goal is to provide the LLM with an enriched context that encompasses both the initial question and the insights from the relevant demonstration examples. The combination of the question and the demonstration examples can be denoted as $\seq{V}\oplus\seq{D}\oplus
\seq{q}$, which is the \textit{prompt} for the LLM. When we input this \textit{prompt} into the LLM, we obtain our final result:

\begin{equation}
\mathrm{Answer} = \lambda(\seq{V}\oplus\seq{D}\oplus\seq{q})
\end{equation}

In this equation, \( \lambda \) represents the prediction process of the LLM. This final prediction step is essential, as it embodies the entire process we've established. It ensures that the LLM takes into account both the original question and the additional information from the demonstration set to generate a well-informed and accurate answer.

\section{Experiments}

\subsection{Experimental Setup}

\paragraph{Datasets}

In experiments, we utilized two benchmark datasets for multi-modal CoT reasoning: 1) \textbf{ScienceQA}~\cite{lu2022learn} is a comprehensive benchmark designed to evaluate multi-modal reasoning ability, particularly for large-scale language models. 2) \textbf{MathVista}~\cite{lu2023mathvista} is designed to evaluate the mathematical reasoning capabilities of LLMs and LMMs in visual contexts.

\paragraph{Models}

In our experiments, we employ ChatGPT~\citep{openai2022chatgpt}, GPT-4~\citep{OpenAI2023GPT4TR} and GPT-4V~\citep{OpenAI2023GPT4TR} via OpenAI API~\footnote{\url{https://platform.openai.com/docs/model-index-for-researchers}}. We utilize\textsc{GPT-3.5-Turbo} for ChatGPT, while \textsc{GPT-4} is employed for GPT-4. As for the evaluation of GPT-4V, we use \textsc{GPT-4-Vision-Preview}.

\begin{table*}[!ht]
    \centering
    \caption{Experimental results on ScienceQA~\citep{lu2022learn}. We evaluated the performance of our system by comparing it with various baseline models, which encompassed both supervised and unsupervised models. Results indicate that our proposed approach \textit{CoT-MM-Retrieval} outperforms previous state-of-the-art models in terms of average accuracy and nearly all question categories on ScienceQA. It should be noted that * represents the best results of our method where we employ more demonstration examples.}
    \resizebox{1.0\linewidth}{!}{
    \begin{tabular}{lclllllllll}
    \toprule
        Methods & Supervised & Avg & NAT & SOC & LAN & TXT & IMG & NO & G1-6 & G7-12 \\ 
    \midrule
        Random Chance & \cmark & 39.8 & 40.3 & 46.1 & 29.3 & 47.5 & 40.1 & 33.7 & 39.4 & 40.7 \\
        LLaVa (GPT-4)~\citep{liu2023llava} & \cmark & 92.5 & 91.6 & 96.7 & 91.1 & 90.6 & 89.0 & 93.5 & 92.7 & 92.2 \\
        LLaMA-SciTune (CTOM)~\citep{horawalavithana2023scitune} & \cmark & 90.0 & 89.3 & 95.6 & 87.0 & 93.1 & 86.7 & 91.8 & 84.4 & 91.3 \\
        \hline
        ChatGPT CoT~\citep{lu2023chameleon} & \xmark & 78.3 & 78.8 & 71.0 & 83.2 & 77.4 & 67.9 & 86.1 & 80.7 & 74.0 \\ 
        Chameleon (ChatGPT)~\citep{lu2023chameleon} & \xmark & 79.9 & 81.6 & 70.6 & \textbf{84.0} & 79.8 & 70.8 & 86.6 & 81.9 & 76.5 \\ 
        ChatGPT CoT-MM-Retrieval & \xmark & 84.7 & 84.4 & 86.4 & 83.8 & 83.1 & 79.2 & \textbf{87.3} & 86.9 & 80.6 \\
        ChatGPT CoT-MM-Retrieval* & \xmark & \textbf{86.4} & \textbf{84.6} & \textbf{93.9} & 83.9 & \textbf{83.4} & \textbf{82.8} & \textbf{87.3} & \textbf{89.1} & \textbf{81.6} \\
        \hline
        GPT-4 CoT~\citep{lu2023chameleon} & \xmark & 84.0 & 85.5 & 72.4 & 90.3 & 82.7 & 71.5 & 92.9 & 86.7 & 79.0 \\
        Chameleon (GPT-4)~\citep{lu2023chameleon} & \xmark & 86.5 & 89.8 & 74.1 & 89.8 & 88.3 & 77.6 & 92.1 & 88.0 & 83.7 \\
        GPT-4 CoT-MM-Retrieval & \xmark & 90.9 & \textbf{90.2} & 89.2 & \textbf{93.5} & 89.3 & 85.1 & \textbf{94.9} & 92.1 & 88.7 \\
        GPT-4 CoT-MM-Retrieval* & \xmark & \textbf{92.5} & \textbf{90.2} & \textbf{97.2} & \textbf{93.5} & \textbf{89.4} & \textbf{88.6} & \textbf{94.9} & \textbf{93.5} & \textbf{90.8} \\
        \hline
        GPT-4V (zero-shot) & \xmark & 90.4 & 92.7 & 83.2 & 91.3 & 91.6 & 85.3 & 92.9 & 91.3 & 88.7 \\
        GPT-4V CoT-MM-Retrieval & \xmark & \textbf{93.1} & \textbf{94.5} & \textbf{89.5} & \textbf{93.1} & \textbf{93.7} & \textbf{89.7} & \textbf{94.5} & \textbf{94.1} & \textbf{91.3} \\
        Human Average~\citep{lu2022learn} & \xmark & 88.4 & 90.2 & 85.0 & 87.5 & 89.6 & 87.5 & 88.1 & 91.6 & 82.4 \\
    \bottomrule
    \end{tabular}
    }
    \label{tab-1:_main_results_scienceqa}
\end{table*}

\paragraph{Implementation Details}

The \textsc{Text-Encoder} and \textsc{Visual-Encoder} used to encode textual and visual context of the CoT example are models pre-trained on large-scale corpora and images. Specifically, for intra-modality similarity, we encode texts and images using SentenceBERT~\citep{reimers2019sentencebert} and ViT~\citep{dosovitskiy2021image}~(ViT-base-patch16-224), respectively. For cross-modality similarity, we encode texts and images using CLIP~\citep{radford2021learning_clip}. Specifically, we followed Chameleon, which involved concatenating meta data and knowledge retrieval results with the current question as our baseline. Our text-based question method is subsequently developed based on this foundation. For the incorporation of visual context, we use BLIP~\cite{li2023blip2} and GPT-4V~\cite{OpenAI2023GPT4TR} to obtain image captions for ScienceQA and MathVista, respectively. We chose the training set of ScienceQA as the demonstration pool. As for MathVista, we evaluated the performance on the test-mini. Given that the test-set does not have published answers and is more substantial in quantity, we chose to use it as the demonstration pool. Since there are no answers or rationale in MathVista's test-set, we used the responses from GPT-4V zero-shot as the basis for our reasoning. To be specific, for GPT-4V MM-Retrieval, our text prompt is the same as LLMs, but we will add the image of the question after the text prompt.

\begin{table*}[!ht]
    \centering
    \caption{Experimental results on MathVista~\citep{lu2023mathvista}. We have evaluated the performance of our retrieval method in comparison with the traditional CoT method. Our proposed approach \textit{CoT-MM-Retrieval} demonstrates exceptional performance across nearly all question categories within MathVista.}
    \resizebox{1.0\linewidth}{!}{
    \begin{tabular}{llllllllllllll}
    \toprule
        Methods & ALL & FQA & GPS & MWP & TQA & VQA & ALG & ARI & GEO & LOG & NUM & SCI & STA \\ 
    \midrule
        Random Chance~\citep{lu2023mathvista} & 17.9 & 15.5 & 24.1 & 4.5 & 23.4 & 24.3 & 25.8 & 13.8 & 22.7 & 13.4 & 8.8 & 15.8 & 14.3 \\
        \hline
        ChatGPT CoT~\citep{lu2023mathvista} & 33.2 & 26.0 & 31.7 & 35.5 & 48.1 & 30.2 & 32.4 & 32.3 & 33.0 & \textbf{16.2} & 17.4 & \textbf{54.9} & 36.2 \\
        ChatGPT CoT-MM-Retrieval & \textbf{41.6} & \textbf{33.1} & \textbf{40.9} & \textbf{50.0} & \textbf{59.5} & \textbf{30.7} & \textbf{44.5} & \textbf{40.8} & \textbf{41.8} & 10.8 & \textbf{18.8} & 54.1 & \textbf{46.2} \\
        \hline
        GPT-4 CoT~\citep{lu2023mathvista} & 33.2 & 27.9 & 31.7 & 31.2 & 51.9 & 28.5 & 33.5 & 30.9 & 32.2 & 13.5 & 12.5 & \textbf{58.2} & 37.9 \\
        GPT-4 PoT~\citep{lu2023mathvista} & 33.9 & 30.1 & 39.4 & 30.6 & 39.9 & 31.3 & 37.4 & 31.7 & 41.0 & 18.9 & \textbf{20.1} & 44.3 & 37.9 \\
        GPT-4 CoT-MM-Retrieval & \textbf{46.8} & \textbf{43.1} & \textbf{45.7} & \textbf{52.7} & \textbf{63.9} & \textbf{32.4} & \textbf{50.2} & \textbf{45.0} & \textbf{46.4} & \textbf{43.2} & 18.1 & 55.7 & \textbf{52.5} \\
        \hline
        GPT-4V (zero-shot) & 49.3 & 42.8 & \textbf{54.8} & 53.8 & 60.1 & 38.6 & \textbf{54.1} & 46.7 & \textbf{54.0} & 29.7 & 25.7 & 60.7 & 54.2 \\
        GPT-4V CoT-MM-Retrieval & 52.0 & 44.6 & 47.1 & 59.7 & \textbf{66.5} & 35.8 & 50.9 & 47.6 & 48.5 & 37.8 & 26.4 & 58.2 & 55.2 \\
        Human Performance~\citep{lu2023mathvista} & \textbf{60.3} & \textbf{59.7} & 48.4 & \textbf{73.0} & 63.2 & \textbf{55.9} & 50.9 & \textbf{59.2} & 51.4 & \textbf{40.7} & \textbf{53.8} & \textbf{64.9} & \textbf{63.9} \\
    \bottomrule
    \end{tabular}
    }
    \label{tab-1:_main_results_mathvista}
\end{table*}

\subsection{Results}

\begin{figure*}
    \centering
  \includegraphics[width=1\linewidth]{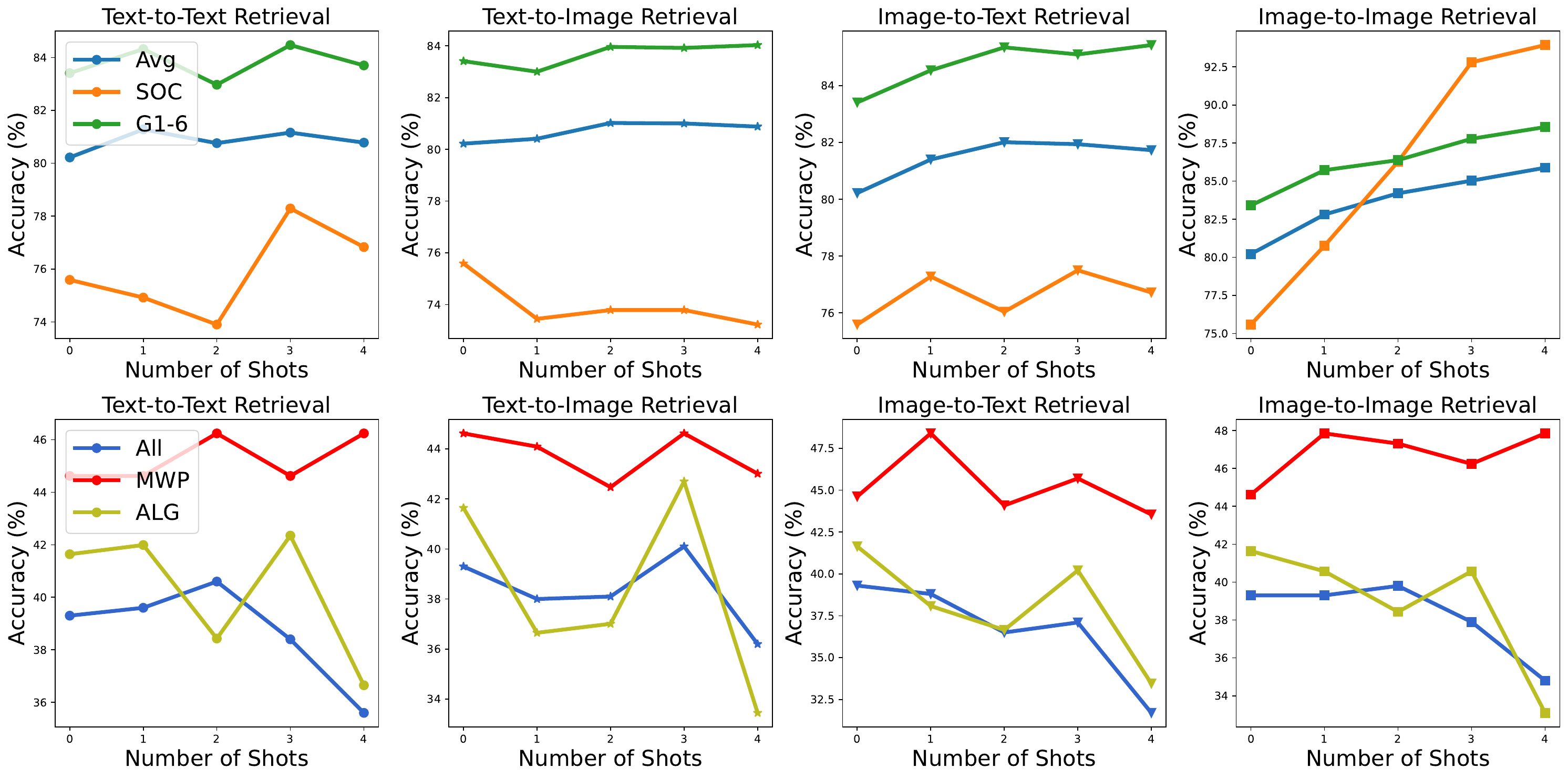}
    \caption{Ablation study of four retrieval methods: \textit{Text-to-Text Retrieval, Text-to-Image Retrieval, Image-to-Text Retrieval, Image-to-Image Retrieval} on ScienceQA~(upper) and MathVista~(bottom). We inspect the performance of each retrieval approach under different amount of demonstration examples.}
    \label{fig:ab-four}
\end{figure*}

\begin{table}[!ht]
    \centering
        \caption{Sampling strategy in ScienceQA and MathVista.}
    \resizebox{1\linewidth}{!}{
    \begin{tabular}{lll}
    \toprule
        Dataset & $q^{v} \notin q$ & $q^{v} \in q$\\ 
    \midrule
        ScienceQA & $S(D_{k4}^{T2I}, D_{k2}^{T2T})$ & $D_{k1}^{I2I}$\\
        MathVista & $\emptyset$ & $S(D_{k2}^{T2T}, D_{k1}^{I2I})$ \\
    \bottomrule
    \end{tabular}
    }
    \label{tab_sampling_strategy}
\end{table}

In our experiments conducted on the ScienceQA dataset~\citep{lu2022learn}, as shown in Table~\ref{tab-1:_main_results_scienceqa}, various models were evaluated for their performance across different question types. The results of baseline models are taken from the ScienceQA leaderboard~\footnote{\url{https://scienceqa.github.io/}}, where we distinguish \textit{supervised} systems and \textit{unsupervised} systems based on whether they are explicitly trained on the training set of ScienceQA. Our approach, termed \textit{CoT-MM-Retrieval}, utilizes two demonstration examples, providing a fair comparison with the \textit{Chameleon}~\citep{lu2023chameleon}, while \textit{CoT-MM-Retrieval*} indicates our best performance with more demonstration examples.
For the models based on ChatGPT: 1) The \textit{Chameleon~(ChatGPT)}~\citep{lu2023chameleon} slightly outperform the base \textit{ChatGPT CoT} with an average of 79.9\%. 2) Our approach \textit{ChatGPT CoT-MM-Retrieval} based on ChatGPT with retrieval augmentation obtained an average accuracy of 84.7\%, outperforming previous state-of-the-art \textit{Chameleon} by 4.8\% 4) Among these, \textit{ChatGPT CoT-MM-Retrieval*} achieved the best performance with an average score of 86.4\%. For the GPT-4 based models: 1) \textit{Chameleon~(GPT-4)}~\citep{lu2023chameleon}, representing the previous state-of-the-art, achieved an average of 86.5\% 2) Our method, \textit{GPT-4 CoT-MM-Retrieval*}, surpassed the \textit{Chameleon~(GPT-4)} by 6\%, achieving an average score of 92.5\%. It set a new state-of-the-art, especially in question types like SOC and NO, with scores of 97.2\% and 94.9\% respectively. For the model based on GPT-4V, our method \textit{GPT-4V CoT-MM-Retrieval} which using the problem image, has surpassed GPT-4V in zero shot's average score by 2.7\%, indicating that our method is applicable not only to LLMs, but also to LMMs.

In our experiments conducted on the MathVista dataset~\citep{lu2023mathvista}, as shown in Table~\ref{tab-1:_main_results_mathvista}. Our approach, termed \textit{CoT-MM-Retrieval}, utilizes two demonstration examples, providing a fair comparison with \textit{CoT} and \textit{PoT}. 
For the models based on ChatGPT, our method \textit{ChatGPT CoT-MM-Retrieval} surpassed the \textit{ChatGPT CoT} by 8.4\%, achieving an average score of 41.6\%. For the models based on GPT-4, our method \textit{GPT-4 CoT-MM-Retrieval} surpassed the \textit{ChatGPT CoT} and \textit{ChatGPT PoT} by 13.6\% and 12.9\% respectively. For the models based on GPT-4V, our method \textit{GPT-4V CoT-MM-Retrieval} surpassed GPT-4V in zero shot average score by 2.7\%, and among the 14 metrics, 8 of them have surpassed zero shot. It is worth noting that in MathVista, due to the difficulty of the math problems, the current GPT-4V cannot surpass humans in average scores.

\subsection{Ablation Studies}

\begin{table*}[ht]
\centering
\caption{Ablation Studies of the three major components of our proposed approach: \textit{Visual Knowledge, Retrieval Mechanism, Stratified Sampling}. We present results on ScienceQA and MathVista where \xmark means we do not employ the corresponding method in our approach and \cmark means we use the corresponding method in the system.} 
\resizebox{0.7\textwidth}{!}{
\begin{tabular}{ccccccccc}
\toprule
    \multirow{2}{*}{Vsion Knowledge} & \multirow{2}{*}{Retrieval} & \multirow{2}{*}{Stratified Sampling} & \multicolumn{3}{c}{ScienceQA} & \multicolumn{3}{c}{MathVista} \\
    \cmidrule(lr){4-6} \cmidrule(lr){7-9}
     & & & Avg & SOC & G1-6 & ALL & MWP & ALG \\
\midrule
\xmark & \xmark & \xmark & 77.8 & 71.2 & 80.5 & 28.0 & 12.4  & 35.9 \\
\cmark & \xmark  & \xmark & 80.7 & 74.6 & 83.7 & 39.4 & 45.7 & 39.9 \\
\cmark & \cmark (T2T)  & \xmark & 80.8 & 73.9 & 83.0 & 40.6 & 46.2 & 38.4 \\
\cmark & \cmark (T2I)  & \xmark & 81.0 & 73.8 & 84.0 & 38.1 & 42.5 & 37.0 \\
\cmark & \cmark (I2T)  & \xmark & 82 & 76 & 85.4 & 36.5 & 44.1 & 36.7 \\
\cmark & \cmark (I2I)  & \xmark & 84.2 & 86.3 & 86.4 & 39.8 & 47.3 & 38.4 \\
\cmark & \cmark  & \cmark & \textbf{84.7} & \textbf{86.4} & \textbf{86.9} & \textbf{41.6} & \textbf{50.0} & \textbf{44.5} \\
\bottomrule
\end{tabular}
}
\label{tab:ablation_studies}
\end{table*}

We conduct analysis towards the effect of different retrieval methods and the amount of demonstration examples~(shots in few-shot learning) in Equation~\ref{eq:mm_retrieval}. The results are shown in Figure~\ref{fig:ab-four}. 

All four retrieval methods: 1) Text-to-Text~(T2T), 2) Text-to-Image~(T2I), 3) Image-to-Text~(I2T), 4) Image-to-Image~(I2I) are explored with increasing shots~(from 0 to 4 for varying $k_{1}, k_{2}, k_{3}, k_{4}$) to study their impact on the model's performance. The performance metrics are provided for various question types, allowing us to discern patterns and variations across different categories. 
The results in Figure~\ref{fig:ab-four} firstly show that adding demonstration examples in the context can improve the overall accuracy especially for ScienceQA and MathVista. We can also observe from Figure~\ref{fig:ab-four}: 1) Text-to-Text Retrieval: The accuracy for T2T retrieval remains fairly consistent as the number of shots increases on ScienceQA. Specifically, the average accuracy ranges between 80.8\% and 81.3\%. And the average accuracy ranges between 35.6\% and 40.6\% on MathVista. The highest accuracy for this method is achieved with 1 shot (81.3\%) and 2 shots(40.6\%) on ScienceQA and MathVista, suggesting that adding more demonstration examples does not always guarantee performance improvement. 2) Text-to-Image Retrieval: The performance is similar to T2T, with average accuracy ranging from 80.4\% to 81\% on ScienceQA and from 36.2\% to 40.1\% on MathVista. For this method, the highest accuracy is achieved with 2 shots (81\%) on ScienceQA, and 3 shots(40.1\%) on MathVista. 3) Image-to-Text Retrieval: The accuracy for this method is slightly more varied than the previous two, ranging from 81.4\% to 82\% on ScienceQA and from 31.7\% 38.8\% on MathVista. Here, 2 shots provide the best average performance at 82\% on ScienceQA, and 1 shot (38.8\%) is the highest accuracy. 4) Image-to-Image Retrieval: This performance is similar to the Text-to-Text Retrieval on MathVista, with average accuracy ranges from 34.8\% to 39.8\%. The highest accuracy for this method is 2 shots(39.8\%). On ScienceQA, this retrieval method showcases the most interesting trend. The accuracy improves significantly with increasing shots, starting from 82.8\% with 1 shot and reaching 85.9\% with 4 shots. On ScienceQA, the G1-6 type consistently performs well, with accuracy usually above 84\%. The choice of retrieval method and the number of shots plays a crucial role in determining model performance.
Our ablation study results demonstrate the robustness and adaptability of our proposed retrieval strategies across different modalities and varying number of shots. Notably, the consistent performance across diverse question types emphasizes the efficacy of our approach in enhancing the reasoning capabilities of LLMs.

\section{Conclusion}

In this paper, we proposed a novel method to address the challenge of selecting suitable demonstration examples for multi-modal reasoning for LLMs~\citep{lu2022learn}. By integrating retrieval mechanisms with LLMs and emphasizing the modality connection between text and images, our approach aims to improve the efficacy of LLMs for multi-modal Chain-of-Thoughts~(CoT)~\citep{cot,zhang2023multicot} reasoning. Furthermore, the introduction of Stratified Sampling in our methodology ensures that LLMs are exposed to a varied and comprehensive set of demonstration examples. In our experiments on the ScienceQA dataset~\citep{lu2022learn} and MathVista dataset~\citep{lu2023mathvista}, our method consistently outperformed existing state-of-the-art models like Chameleon~\citep{lu2023chameleon} and ChatGPT PoT~\citep{lu2023mathvista}. These empirical results validate our hypothesis that combining LLMs with tailored retrieval mechanisms, like the ones we propose, can significantly enhance multi-modal reasoning performance. As CoT techniques continue to gain traction in the NLP community, our work underscores the importance of effective demonstration example selection. 

Future research should focus on refining the retrieval processes and extending the methodologies developed in this study to additional multi-modal tasks. These tasks could include those where the outputs from Large Language Models (LLMs) encompass multiple modalities, such as text-to-image and text-to-video generation~\citep{liu2023cultural, wang2023gpt4video}. Additionally, application in specialized domains, like the medical field~\citep{li2023comprehensive_medical}, presents a promising direction. Concurrently, in the development of increasingly sophisticated multi-modal LLMs, particularly those employing CoT reasoning, addressing the issue of hallucination is important~\citep{ji2023survey,zhang2023sirens}. We believe that our work lays a strong foundation for these future explorations.

\section*{Limitations}

In this paper, we present a new approach for augmenting LLMs with multi-modal retrieval for CoT examples. However, there are a few limitations for our work. Firstly, we only test our approach on two datasets: ScienceQA and MathVista. These two datasets are mainly about science and math topics which do not have a broad coverage of other complex reasoning tasks. Therefore, our approach should also be evaluated on other complex reasoning tasks. Secondly, due to the limitation of resources we only conduct experiments on close-source systems and do not carry out experiments on open-source LLMs, making it harder and more expensive to reproduce our results. Moreover, due to the nature of these close-source LLMs we can not fully eliminate the risk of data contamination. Therefore, our approach should also be evaluated comprehensively on more representative languages.

\bibliography{custom}

\clearpage
\appendix

\section{Appendices}

\subsection{Case Study}
\begin{figure*}[t]
  \centering
    \includegraphics[width=1\textwidth]{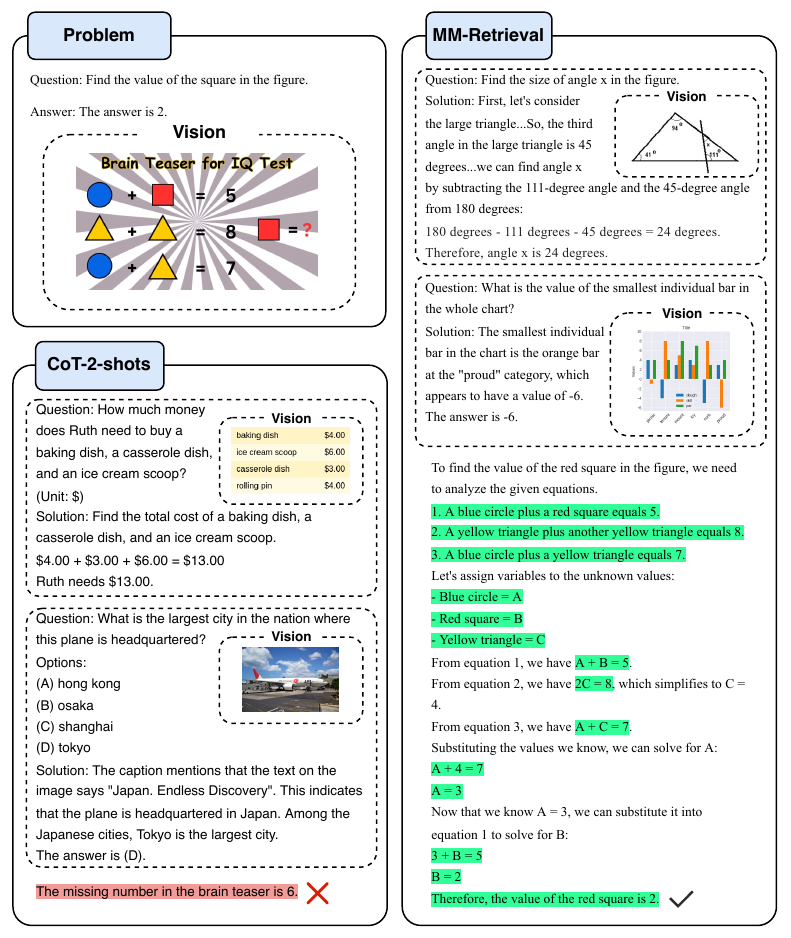}
  \caption{An example of our method on MathVista. The left side depicts the common 2-shot CoT approach for LLM, while the right side showcases our MM-Retrieval method retrieving demonstration examples as relevant as possible to enhance the reasoning ability of LLM.}
  \label{fig:case_mathvista_98}
\end{figure*}

\begin{figure*}[t]
  \centering
    \includegraphics[width=1\textwidth]{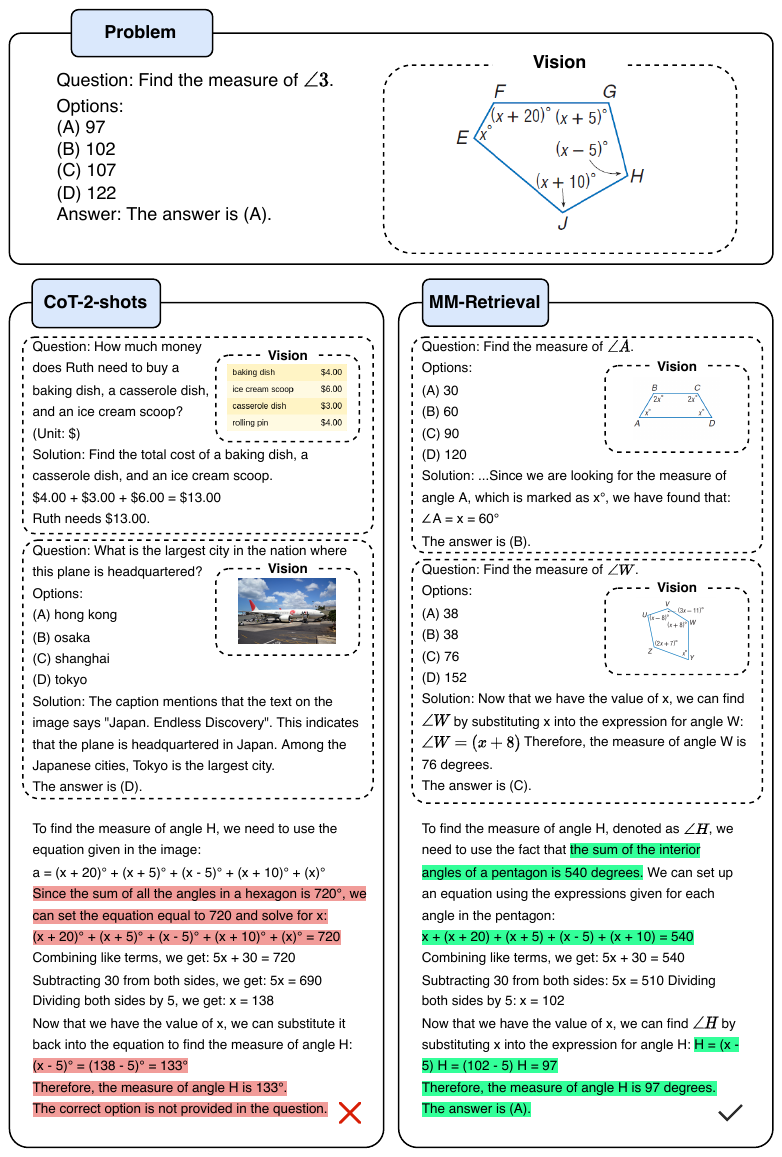}
  \caption{An example of our method on MathVista. The left side illustrates the conventional 2-shot CoT technique used for LLM, while the right side highlights our MM-Retrieval method, which aims to retrieve demonstration examples that are as relevant as possible to bolster the reasoning capabilities of LLM.}
  \label{fig:case_mathvista_1}
\end{figure*}

\begin{figure*}[!t]
    \centering
    \includegraphics[width=0.96\linewidth]{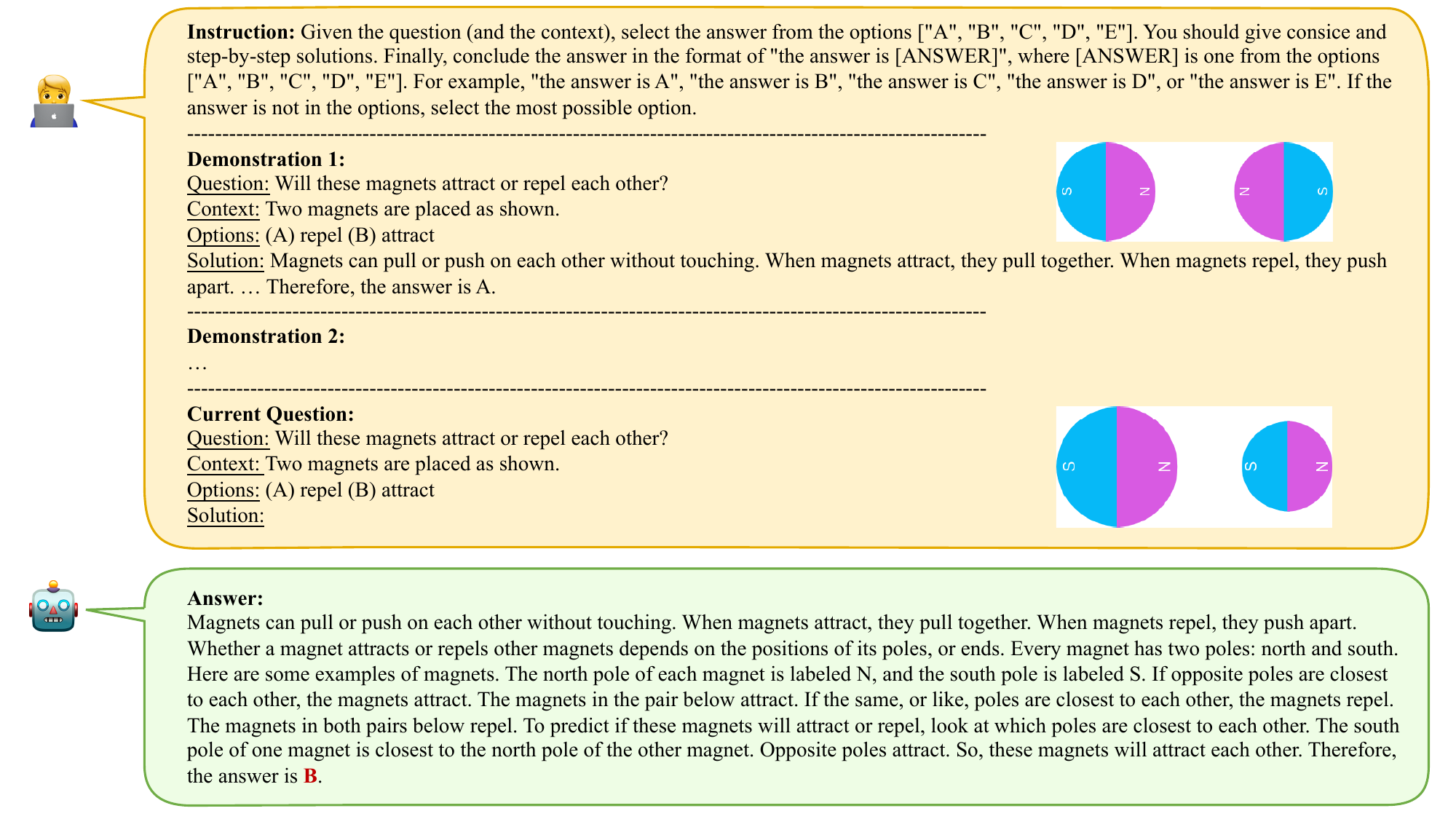}
    \caption{An example of our method on ScienceQA. For the question with visual context, our approach retrieves demonstration examples as relevant as possible to enhance the reasoning ability of LLM.}
    \label{fig:case1}

    \includegraphics[width=0.96\linewidth]{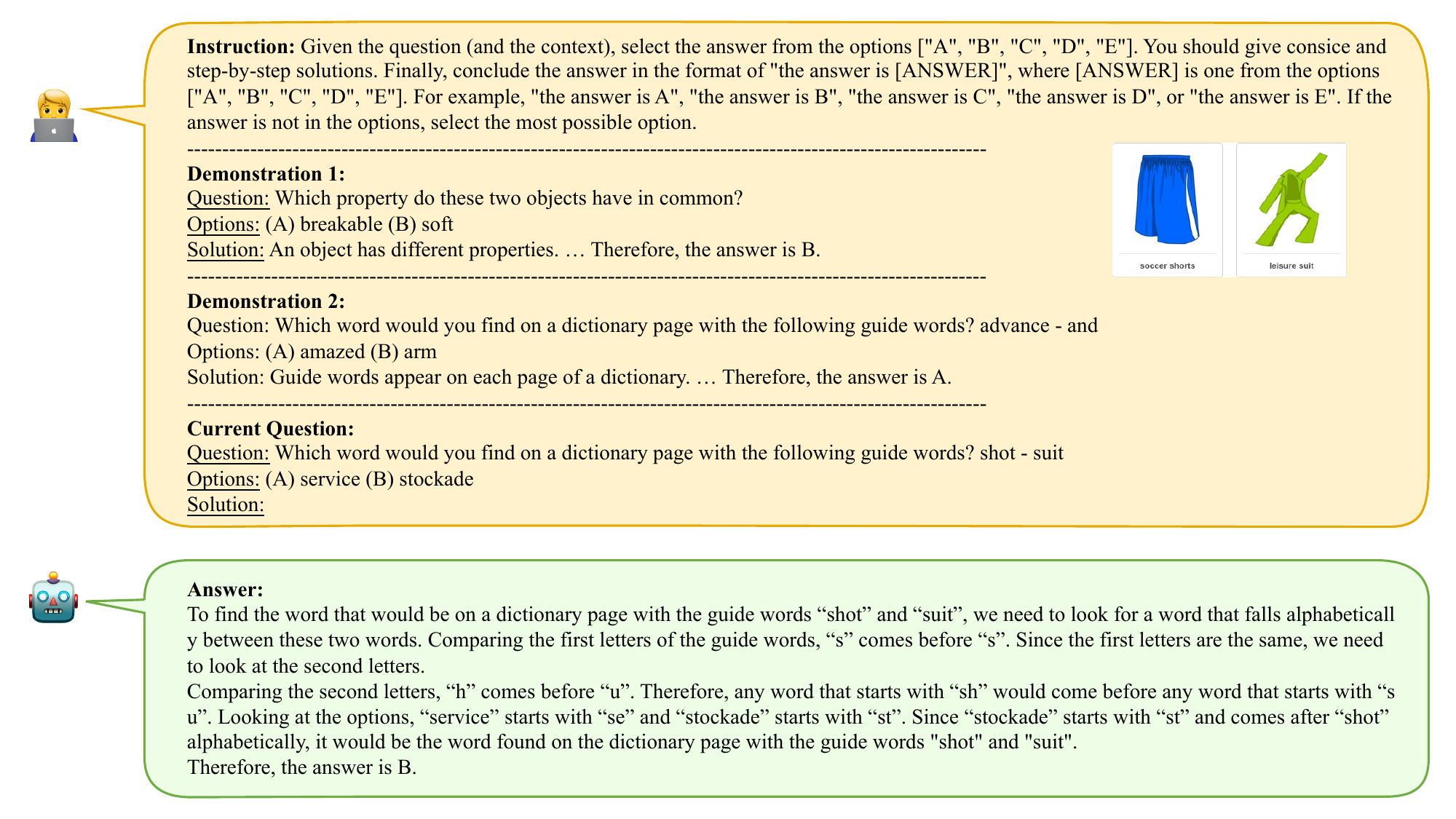}
    \caption{An example of our method on ScienceQA. For the question without visual context, our method retrieves demonstration examples with both visual context and textual context, enhancing the diversity of demonstration examples to improve the reasoning ability of LLM.}
    \label{fig:case2}
\end{figure*}

We present a case study to demonstrate the effectiveness of our approach in Figure~\ref{fig:case_mathvista_98}, Figure~\ref{fig:case_mathvista_1}, Figure~\ref{fig:case1} and Figure~\ref{fig:case2}. 

In Figure~\ref{fig:case_mathvista_98} and Figure~\ref{fig:case_mathvista_1}, compared to CoT-2-shots that uses fixed samples regardless of the question, our method is able to retrieve Demonstrations based on the current question, and control the diversity of samples through hierarchical sampling, thus effectively stimulating the reasoning ability of LLMs.

In Figure~\ref{fig:case1}, we illustrate our approach's capability when dealing with a question with an image, such as determining the interaction between two magnets. Our approach enhances the LLM's reasoning by retrieving related questions that provide deeper understanding of magnetic properties, such as attraction and repulsion based on pole alignment, enriching the LLM's knowledge base for more accurate inference. 

Figure~\ref{fig:case2} highlights our approach's adaptability to questions without visual context, such as deducing alphabetical order in a dictionary. It showcases the approach's use of varied demonstrations—incorporating both visual and non-visual information—to strengthen the LLM's ability to discern patterns and sort information. This demonstrates the approach's versatility in improving reasoning by broadening the range of demonstrations, which in turn equips the LLM to handle a diverse set of textual queries effectively.

\end{document}